\documentclass[lettersize,journal]{IEEEtran}
\usepackage{amsmath,amsfonts}
\usepackage{algorithmic}
\usepackage{algorithm}
\usepackage{array}
\usepackage[caption=false,font=normalsize,labelfont=sf,textfont=sf]{subfig}
\usepackage{textcomp}
\usepackage{stfloats}
\usepackage{url}
\usepackage{verbatim}
\usepackage{graphicx}
\usepackage{cite}
\usepackage{multirow}
\usepackage{color}

\begin{document}

\title{Nowhere to Hide: A Lightweight Unsupervised Detector against Adversarial Examples}
\author{Hui Liu, Bo Zhao, Kehuan Zhang~\IEEEmembership{Member,~IEEE,} and Peng Liu~\IEEEmembership{Member,~IEEE,}
\thanks{Hui Liu, Bo Zhao are with the School of Cyber Science and Engineering, Wuhan University, Wuhan, 430072 China e-mail: zhaobo@whu.edu.cn.}

\thanks{Kehuan Zhang is with the College of Information Engineering, Chinese University of Hong Kong, HK, 999077 China e-mail: khzhang@ie.cuhk.edu.hk.}

\thanks{Peng Liu is with the College of Information Sciences and Technology, Pennsylvania State University, PA, 16801 US e-mail: pliu@ist.psu.edu.}}



\maketitle

\begin{abstract}
Although deep neural networks (DNNs) have shown impressive performance on many perceptual tasks, they are vulnerable to \emph{adversarial examples} that are generated by adding slight but maliciously crafted perturbations to benign images. Adversarial detection is an important technique for identifying adversarial examples before they are entered into target DNNs. Previous studies to detect adversarial examples either targeted specific attacks or required expensive computation. How design a lightweight unsupervised detector is still a challenging problem. In this paper, we propose an \textbf{A}uto\textbf{E}ncoder-based \textbf{A}dversarial \textbf{E}xamples (\textbf{AEAE}) detector, that can guard DNN models by detecting adversarial examples with low computation in an unsupervised manner. The \textbf{AEAE} includes only a shallow autoencoder but plays two roles. First, a well-trained autoencoder has learned the manifold of benign examples. This autoencoder can produce a large reconstruction error for adversarial images with large perturbations, so we can detect significantly perturbed adversarial examples based on the reconstruction error. Second, the autoencoder can filter out the small noise and change the DNN's prediction on adversarial examples with small perturbations. It helps to detect slightly perturbed adversarial examples based on the prediction distance. To cover these two cases, we utilize the reconstruction error and prediction distance from benign images to construct a two-tuple feature set and train an adversarial detector using the isolation forest algorithm. We show empirically that the \textbf{AEAE} is unsupervised and inexpensive against the most state-of-the-art attacks. Through the detection in these two cases, there is nowhere to hide adversarial examples.
\end{abstract}


\section{Introduction}

\IEEEPARstart{D}{eep}  neural networks (DNNs) \cite{Huang18} have been widely used in various fields, and achieved impressive performance on many intelligent tasks, such as biometric identification \cite{BlancoG19, WuD20}, malware detection\cite{DingMGSCZ21}, etc. However, a large number of studies have shown that attackers can generate adversarial examples to deceive the well-trained DNN model. Adversarial attacks have potentially disastrous consequences for DNNs-based systems. In the physical world, attackers can project well-crafted perturbations onto real-world objects, transforming them into adversarial examples. \cite{Lovisotto21}.

How to effectively detect adversarial examples has been a challenging task. During the last few years, researchers have made great efforts to design state-of-the-art adversarial detectors, e.g., MagNet \cite{Meng17}, feature squeezing \cite{Xu18}, NIC \cite{MaLTL019}, DLA \cite{SperlKCLB20}. However, these approaches do not always yield satisfactory results. For example, the accuracy of MagNet and feature squeezing needs to be further improved against some specific attacks \cite{Meng17, Xu18}. NIC \cite{MaLTL019} has to train a derived model for each layer of the target DNN, which cause expensive computation cost. DLA \cite{SperlKCLB20} assumes a specific process for generating adversarial examples, so its accuracy would decrease when new attacks are launched. These concrete limitations motivate us to design an effective and lightweight adversarial detector in an unsupervised manner.

We observe that the effect of adversarial detection is sensitive to the perturbation level. Specifically, the attacker can escape adversarial detection by adjusting the perturbation level, such as changing the control parameters of adversarial perturbations or replacing a new attack method. For example, MagNet \cite{Meng17} can detect the FGSM attack with control parameter $\epsilon=32$ in 100\% detection rate, but when the control parameter $\epsilon=16$, the detection rate of MagNet is only 3\%. Feature-filter \cite{LiuH22} is very effective against C\&W attacks \cite{Carlini017}, but not against BIM attacks \cite{KurakinGB17a}.

Based on the key observation, we divide adversarial detection into two sub-tasks according to the perturbation level, that is, adversarial detection with significant perturbations and slight perturbations. Meng et al. \cite{Meng17} proposed that the autoencoder could learn an approximate manifold of benign examples and the detector based on reconstruction error was effective in detecting significantly perturbed adversarial examples. In our recent study \cite{LiuH22}, we revealed that feature filtering contributes significantly to the detection of slightly perturbed adversarial examples. And the autoencoder is an effective feature-filtering method. These findings inspire us to design an autoencoder-based detector that simultaneously detects adversarial examples with both high and low perturbation levels.

In this paper, we design an \textbf{AEAE} detector that is constructed by a shallow autoencoder. This autoencoder is trained only on benign examples. If an input is drawn from the benign dataset, the autoencoder produces a small reconstruction error. Otherwise, if an input is a significantly perturbed adversarial example, the autoencoder produces a larger reconstruction error. Hence, we use reconstruction error to estimate how far a test example is from the manifold of benign examples. To detect slightly perturbed adversarial examples, \textbf{AEAE} inputs the original image and its reconstructed version into the target DNN model. Since the adversarial example is more sensitive to filtering than the benign example, there is a larger DNN's prediction distance between the adversarial example and its reconstructed version. We use the target DNN's prediction distance to estimate the sensitivity of the original image to the autoencoder. In order to improve the generalization of our detector, we use reconstruction error and prediction distance from benign examples to construct a two-tuple feature set. This feature set is employed to train an outlier detector based on the isolation forest algorithm. If the outlier detector decides a two-tuple feature of an original image belongs to a different distribution, this image is considered to be adversarial; otherwise, it is considered benign.

In summary, this paper makes the following contributions:
\begin{itemize}
\item We observe that the effect of adversarial detection is sensitive to the perturbation level. Thus, we argue that adversarial detection should be divided into two independent tasks according to the perturbation level: detecting adversarial examples with both significant and slight perturbations.

\item We only train one very shallow autoencoder on benign examples, which can simultaneously find intrinsic features from both significantly and slightly perturbed adversarial examples.

\item We propose an effective adversarial detector that is referred to as \textbf{AEAE}. The \textbf{AEAE} neither targets specific adversarial attacks nor modifies neural networks. Thus, the \textbf{AEAE} is an unsupervised and inexpensive adversarial detector.
\end{itemize}

The remainder of this paper is organized as follows: We introduce the related work in Section \uppercase\expandafter{\romannumeral2} and the preliminaries in Section \uppercase\expandafter{\romannumeral3}. We present the detailed design of the \textbf{AEAE} in Section \uppercase\expandafter{\romannumeral4}, followed by the experimental results in Section \uppercase\expandafter{\romannumeral5}. Finally, we conclude the paper in Section \uppercase\expandafter{\romannumeral6}.

\section{Related Work}
Adversarial detection, which distinguishes adversarial examples from benign examples, is fundamental for a robust DNN-based system. The last few years have witnessed improvements and developments in adversarial detection. A variety of adversarial detectors are proposed by the research community. Based on different strategies, adversarial detection can be categorized into three group: sample statistics \cite{CohenSG20, Meng17}, auxiliary classifier \cite{MaLTL019, DengYX0021, SperlKCLB20} and input transformation \cite{Xu18, LiuH22, TianS18}.

The detection strategy of sample statistics is based on the assumption that there are different statistical properties between benign examples and adversarial examples. Cohen et al. \cite{CohenSG20} conjectured that DNN's decision boundary might be related to the training data, and adversarial examples could break this relationship. They utilized nearest neighbor influence functions to measure the statistical distance and trained a logistic regression model on these statistical distances as a detector. Meng et al. \cite{Meng17} utilized the reconstruction error of the autoencoder to estimate the statistical distance between the test example and the benign example manifold. This approach is very effective to detect adversarial examples far from the boundary of the manifold. However, some state-of-the-art attacks can generate adversarial examples with very low perturbations, making it difficult to locate the significant statistical properties. Therefore, the detector based on sample statistics seems unlikely to be effective against adversarial examples with slight perturbations.

The auxiliary classifier strategy treats adversarial detection as a binary task. By monitoring benign and adversarial example behaviors, an auxiliary classifier is built to identify them. Ma et al. \cite{MaLTL019} argued that adversarial attacks changed the provenance of layers or the activation value distribution. They trained a derived model for each layer to describe the distribution of the provenance invariant and the value invariant. When both invariants of the original image do not fit distributions of benign examples, this image is determined to be an adversarial example. Given a pre-trained DNN, LiBRe \cite{DengYX0021} converts its last few layers to be Bayesian, in the spirit of leveraging Bayesian neural networks for adversarial detection. Sperl et al. \cite{SperlKCLB20} observed that adversarial examples provoke the dense layer neuron coverage to behave in a unique pattern. They proposed a general end-to-end method DLA to detect adversarial examples. DLA is an alarm model trained on benign and adversarial features. Since DLA assumes some specific adversarial attacks, it seems unlikely to be effective to detect new types of attacks.

The input transformation strategy is based on a key observation: adversarial examples are more sensitive to transformations than benign examples. The basic idea is to measure the prediction inconsistency in a test input and its transformed version. Feature squeezing \cite{Xu18} provides two simple types of squeezing, e.g. bit depth reduction and spatial smoothing. This detector compares the target DNN's prediction on the original image with that on squeezed images to achieve high accuracy and few false positives. Similarly, Liu et al. \cite{LiuH22} explained why the imperceptible adversarial example exists and proposed a feature-filter to further improve performance on adversarial examples with slight perturbations. Tian et al. \cite{TianS18} exploited a set of rotations to yield several transformed versions, and then collected the target DNN's prediction on them. These predicted results are employed to train an alarm model. They claimed that this alarm model could effectively thwart C\&W attacks \cite{Carlini017}. The input transformation strategy should follow a basic criterion, that is, these transformations should be slight enough not to change the classification results of benign examples. This criterion limits the detection performance on significantly perturbed adversarial examples.

Since these detection strategies show significant differences in detecting adversarial examples with different perturbation levels, we divide adversarial detection into two independent tasks, i.e., detecting significantly and slightly perturbed adversarial examples. In the task of detecting significantly perturbed adversarial examples, we take advantage of the autoencoder's capability to learn statistical properties from the benign example manifold. The reconstruction error is employed to measure the distance between an original image and this manifold to detect adversarial examples with significant perturbations. To detect slightly perturbed adversarial examples, we give full play to the denoising capability of the autoencoder to filter out small adversarial perturbations. To cover both types of adversarial examples with different perturbation levels, we construct an auxiliary classifier based on the isolation forest algorithm. This classifier can detect adversarial examples without targeting specific attacks. Combining these three detection strategies, one autoencoder plays two roles and has outstanding generalization.

\section{Preliminaries}
\subsection{Adversarial Attacks}
Consider a classifier $f(x): \mathbb{R}^{d} \rightarrow\{1 \ldots k\}$ to map an input image $x \in \mathbb{X}$ to a label set $\mathcal{C}$ with $k$ classes, where $d$ is the input dimension. The goal of an attacker is to find a perturbation $\delta \in \mathbb{R}^{d}$ to maximize the loss function, e.g., cross-entropy loss $\mathcal{L}_{ce}$, so that $f(x+\delta) \neq f(x)$, where $\delta$ is estimated as
\begin{equation}
\delta^{*}:=\underset{|\delta|_{p} \leq \epsilon}{\arg \max } \mathcal{L}_{c e}(x+\delta, y)
\label{lp}
\end{equation}
where $y$ is the label of $x$. $|\delta|_{p}$ denotes $p$-norm distance, which is used to measure the perturbation level. $p$ can commonly be 0, 2 and $\infty$. The $L_0$ norm measures the number of pixels perturbed in an image. $L_2$ measures the Euclidean distance. The $L_{\infty}$ norm denotes the maximum for all vector elements $\left|\delta_{i}\right|:\rVert\delta\rVert_{\infty}=max(\left|\delta_{i}\right|)$. In what follows, we introduce state-of-the-art adversarial attacks \cite{GoodfellowSS14, KurakinGB17a, MadryMSTV18, Moosavi-Dezfooli16, Carlini017} that generate such perturbations $\delta$.

\subsubsection{Fast Gradient Sign Method}
The fast gradient sign method (FGSM) \cite{GoodfellowSS14} does not require an iterative procedure and only computes a one-step gradient along the direction of the sign of gradient at each pixel. Thus, it is a fast approach using back-propagation to generate adversarial examples. The FGSM attack can be formulated as follows,
\begin{equation}
x'=x + \epsilon \cdot \operatorname{sign}\left(\nabla_{x} J(f(x), y)\right)
\label{fgsm}
\end{equation}
where $\epsilon$ controls the perturbation level. Larger $\epsilon$ indicates greater perturbation. $J(\cdot,\cdot)$ is the loss function, and $y$ is the ground-truth label for $x$.

\subsubsection{Basic Iterative Method}
Kurakin et al. \cite{KurakinGB17a} proposed a basic iterative method (BIM) attack and performed it on the DNN-based system in physical world scenarios. They extended the FGSM by running a small step size for multiple iterations. In each iteration, the BIM attack clips pixel values of intermediate results to ensure that they are in an $\epsilon$-neighborhood of the original image $x$.
\begin{equation}
\begin{aligned}
x'_{0} & = x, \\
x'_{n+1} & = \operatorname{Clip}_{x, \epsilon}\left\{x'_{n}+\alpha \operatorname{sign}\left(\nabla_{x} J\left(x'_{n}, y\right)\right)\right\}
\end{aligned}
\label{bim}
\end{equation}
where $n$ is the number of iterations, $\epsilon$ controls the perturbation level on each step. $\alpha$ is set to 1 in \cite{KurakinGB17a}, meaning that the value of each pixel changes only by 1 on each step. $\operatorname{Clip}_{x, \epsilon}(\cdot)$ performs per-pixel clipping to keep the result in the $L_{\infty}$ $\epsilon$-neighborhood of $x$.

\subsubsection{Projected Gradient Descent}
Different from FGSM, which can be interpreted as a one-step scheme, projected gradient descent (PGD) is a multi-step variant for maximizing the loss function. Madry et al. \cite{MadryMSTV18} applied PGD in a new adversarial attack method defined as follows,
\begin{equation}
x'_{n+1}=\Pi_{x+\mathcal{S}}\left(x'_{n}+\alpha \operatorname{sign}\left(\nabla_{x} J\left(x'_{n}, y\right)\right)\right)
\label{pgd}
\end{equation}
where each run starts at a uniformly random point in the $L_{\infty}$ $\epsilon$-neighborhood example. PGD iteratively updates the perturbation by taking a small step $\alpha = 0.01$ and constraints the total perturbation to $\epsilon$ after each iteration.

\subsubsection{DeepFool}
For an affine classifier $f(x)=\omega^T x+b$, its affine hyperplane is $\Gamma=\left\{x: \omega^T x+b=0\right\}$. The minimal perturbation to change the classifier's decision corresponds to the orthogonal projection of the example $x$ onto the hyperplane $\Gamma$. The perturbation of an affine classifier $f$ can be $\delta^*(x)=-\left(f(x) /\|\omega\|^2\right) \omega$. Based on this geometry concept, Moosavi-Dezfooli et al. \cite{Moosavi-Dezfooli16} proposed DeepFool to search for the minimal perturbation by considering $f$ that is linearized around $x_i$ at each iteration. The minimal perturbation is computed as follows,
\begin{equation}
\begin{gathered}
\underset{\delta_i}{\arg \min }\left\|\delta_i\right\|_2 \\
s.t. \quad f\left(\boldsymbol{x}_i\right)+\nabla f\left(\boldsymbol{x}_i\right)^T \boldsymbol{r}_i=0
\end{gathered}
\end{equation}
where these perturbations $\delta_i$ are accumulated to get the final perturbation $\delta$. By searching within this polyhedron for minimal perturbation, DeepFool can change the classifier's decision and achieve less perturbation than the FGSM attack.

\subsubsection{Carlini/Wagner}
Carlini/Wagner (C\&W) attack \cite{Carlini017} is an iterative approach to generate adversarial with small perturbations. It can be targeted or untargeted for all three ($L_0$, $L_2$ and $L_{\infty}$) norms. The C\&W attack achieves the powerful attack ability, which can generate adversarial examples with small perturbation. Carlini et al. claimed that the $L_2$ attack escaped the adversarial detector better than the other two attacks. The C\&W attack with the $L_2$ norm can be formulated as follows,
\begin{equation}
\begin{aligned}
{\rm min} \quad & \lVert \delta \rVert _2 + c \cdot g(x') \\
s.t. \quad & x'=x+\delta \in \mathbb{R}^{d}\\
\end{aligned}
\label{eq3}
\end{equation}
where $c$ is a suitable constant, and the penalty function \emph{g} is defined as
\begin{equation}
g(x')={\rm max}({\rm max}\{Z(x')_i : i \neq t\} - Z(x')_t, -k)
\label{eq4}
\end{equation}
where $g\left(x^{\prime}\right) \geq 0$ if and only if $f\left(x^{\prime}\right)=l^{\prime}$. In this way, the distance and the penalty term can be better optimized. $Z(x)$ denotes the softmax function and a constant $k$ encourages the solver to find an adversarial example $x^\prime$ that will be classified as label $t$ with high confidence.

\subsection{Autoencoder}
An autoencoder is a type of neural network, which learns a representation for training data and reconstructs the input from this representation. An autoencoder $a e=d \circ e$ consists of an encoder and a decoder. The encoder $e: \mathbb{R}^d \rightarrow \mathbb{H}^n$ and the decoder $d: \mathbb{H}^n \rightarrow \mathbb{R}^d$, where $\mathbb{R}^d$ is the input space of an image and $\mathbb{H}^n$ is a generally lower-dimensional space of latent representation. We train an autoencoder to minimize the reconstruction error, e.g., mean squared error ($MSE$). Therefore, the $MSE$ value is calculated as follows,
\begin{equation}
MSE(\mathbb{X})=\sum_{x \in \mathbb{X}}\|x-a e(x)\|^{2}
\label{eq:autoencoder}
\end{equation}
where $\mathbb{X} \subset \mathbb{R}^{d}$ denotes the training set composed of benign examples. Intuitively, the reconstruction error describes a distance of a given image from the benign example manifold.

On the one hand, our autoencoder only learns features of benign examples. When a significantly perturbed adversarial example is entered into this well-trained autoencoder, it tends to produce a large reconstruction error. On the other hand, the output of the autoencoder is regenerated from latent representation, so the autoencoder could filter out some insignificant feature (noise) of the input, e.g., small adversarial perturbations. It helps us to detect slightly perturbed adversarial examples.

\subsection{Isolation Forest}
The isolation forest \cite{LiuTZ12, TokovarovK22} is an outlier detection approach purely based on the concept of isolation. Anomalies are “few and different", and therefore more susceptible to isolation mechanisms. Liu et al. \cite{LiuTZ12} constructed a binary tree structure called isolation tree ($i$Tree) to effectively isolation instances. Due to the susceptibility to isolation, anomalies are more likely to be isolated closer to the root of an $i$Tree, while normal points are more likely to be isolated at the deeper end of an $i$Tree. The isolation forest builds an ensemble of $i$Trees for a data set. Assume the path length between each data $x$ and the root node is $h(x)$, and the anomaly value is calculated as follows,
\begin{equation}
\begin{gathered}
s(x, n)=2^{\left(-\frac{E(h(x))}{c(n)}\right)} \\
c(n)=2 H(n-1)-\frac{2(n-1)}{n}, H(k)=\ln (k)+\xi
\label{iforest}
\end{gathered}
\end{equation}
where $\xi = 0.57772156649$, $E(h(x))$ is the average of $h(x)$ from a collection of $i$Trees. $s(x,n)$ is the anomaly value of data $x$ in the $n$ samples of a data set. $s(x,n)$ ranges from 0 to 1. A larger $s(x,n)$ value means a higher probability of an outlier.

The isolation forest is an algorithm with a low linear time complexity and a small memory requirement \cite{LiuTZ12}. It can be trained with or without anomalies in the training data, and provide robust detection results. In our approach, the isolation forest is trained only on a feature set from benign examples and does not require any adversarial examples.

\section{Design}
\begin{figure}[htb]
\centering
\includegraphics[height=3.8in]{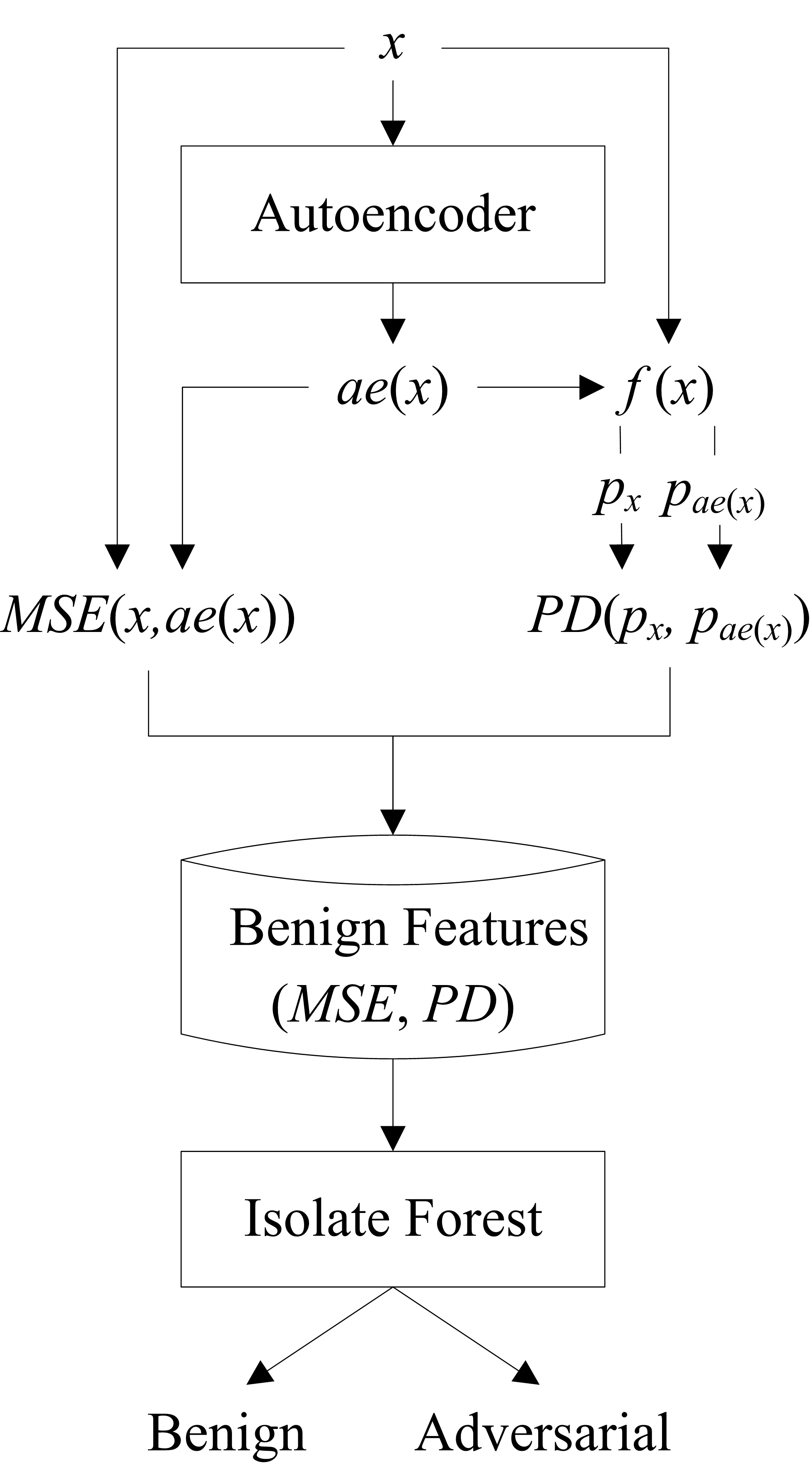}
\caption{Overview of the \textbf{AEAE} detector. An autoencoder plays two roles.}
\label{overview}
\end{figure}

We observe that adversarial detection is susceptible to perturbation levels. The sample statistic strategy is effective for detecting significantly perturbed adversarial examples and the input transformation strategy is effective for detecting slightly perturbed adversarial examples. An autoencoder can play both roles. In the first role, the autoencoder learns statistical features of benign examples to distinguish adversarial examples with significant perturbations. In another role, the autoencoder is capable of filtering out small noises, as so to detect adversarial examples with slight perturbations. Therefore, the design of our detector \textbf{AEAE} considers both types of adversarial examples with different perturbation levels. Figure~\ref{overview} shows the overview of the \textbf{AEAE} based on the autoencoder. The details of the \textbf{AEAE} detector are listed below.

(1) We construct and train an autoencoder to minimize the reconstruction error $MSE$, which is defined in Equation \ref{eq:autoencoder}. For CIFAR-10 and ImageNet datasets, we respectively construct a shallow autoencoder whose details are listed in Table~\ref{tab_autoencoder}. To learn the statistical features of benign examples, this autoencoder is trained only on benign examples $\mathbb{X}$. The training parameters of these autoencoders are listed in Table~\ref{tab_autoencoderParam}.
\begin{table}[htb]
\caption{Autoencoder architecture.}
\centering
\begin{tabular}{|c|c|c|}
\hline
\multirow{2}*{Type}  & \multicolumn{2}{c|}{Output shape} \\ \cline{2-3}
                     & CIFAR-10       & ImageNet  \\
\hline
Input  & $32 \times 32 \times 3$ &  $224 \times 224 \times 3$ \\
\hline
Conv2D.ReLU   &  $3 \times 3 \times 32 $ &  $3 \times 3 \times 64$ \\
\hline
Max pooling   & $2 \times 2$   &  $2 \times 2$ \\
\hline
Conv2D.ReLU   &  $3 \times 3 \times 32 $ &  $3 \times 3 \times 64$ \\
\hline
Up sampling   & $2 \times 2$  &  $2 \times 2$ \\
\hline
Conv2D.Sigmoid &  $3 \times 3 \times 3$   &  $3 \times 3 \times 3$ \\
\hline
\end{tabular}
\label{tab_autoencoder}
\end{table}

\begin{table}[htb]
\caption{Training parameters.}
\centering
\begin{tabular}{|c|c|c|}
\hline
Parameters & CIFAR-10 & ImageNet  \\
\hline
Optimizer  & Adam & Adam  \\
\hline
Learning rate & 0.01 &  0.01 \\
\hline
Loss function & MSE & MSE \\
\hline
mode    & Min  &  Min \\
\hline
Batch size & 64 & 32 \\
\hline
Epochs & 50 & 20 \\
\hline
\end{tabular}
\label{tab_autoencoderParam}
\end{table}

(2) Each benign image $x \in \mathbb{X}$ is entered into the well-trained autoencoder and its reconstructed version $ae(x)$ is generated. We calculate the reconstruction error $MSE$ between $x$ and $ae(x)$.

(3) Both $x$ and $ae(x)$ are entered into the target DNN classifier $f(\cdot)$. This classifier produces prediction vectors $p_x$ and $p_{ae(x)}$ for both images. For CIFAR-10 images, we utilize Kullback–Leibler ($KL$) divergence~\cite{Bahat19} to measure the distance $PD$ between two prediction vectors. The definition of $KL$ is formalized as Equation ~\ref{kl}.
\begin{equation}
KL(p_x||p_{ae(x)}) = \sum p_x log \frac{p_x}{p_{ae(x)}}
\label{kl}
\end{equation}
where $KL$ is a measure of how a prediction distribution $p_x$ of the image $x$ is different from $p_{ae(x)}$ of its reconstructed version $ae(x)$. The $KL$ value is the expectation of the logarithmic difference between the probabilities $p_x$ and $p_{ae(x)}$.

Since ImageNet is a high-resolution dataset with 1,000 classes, the length of the DNN prediction vector $p_x$ for its images is 1,000. The length is so long that the distance $PD$ between the two prediction vectors is not obvious by using $KL$ divergence. Thus, for ImageNet images, we propose the prediction distance by comparing DNN's prediction labels on $x$ and $ae(x)$. The prediction distance $PD$ for ImageNet is formalized as Equation ~\ref{pd}.
\begin{equation}
PD(x)=
\begin{cases}
1, & argmax(f(x)) \neq argmax(f(a e(x))) \\
0, & argmax(f(x))   =  argmax(f(a e(x)))
\end{cases}
\label{pd}
\end{equation}
where $argmax(\cdot)$ returns the indices of the maximum values. When the autoencoder changes the prediction label of the input, the prediction distance $PD$ is set to 1; Otherwise, $PD$ is set to 0.

(4) We construct a two-tuple feature $bf = (MSE,PD)$ for each benign image and obtain a feature set $D_{bf}$. By training an isolation forest algorithm, we obtain an alarm model to distinguish between benign examples and adversarial examples.

Before an original image is entered into the target DNN classifier, this image is input into the autoencoder to yield its reconstructed image. Then, a two-tuple feature of this image is generated by calculating $MSE$ and $PD$ values between this image and its reconstructed version. Finally, the well-trained isolation forest model accepts this two-tuple feature and outputs a predictive result. If the two-tuple feature is an outlier, this image is considered adversarial and the alarm model rejects it; otherwise, it is considered benign and the target DNN classifier gives a prediction label.

The \textbf{AEAE} detector includes two training processes. One is to train an autoencoder on benign examples and the other is to train an isolation forest model on benign features. The \textbf{AEAE} does not assume a specific process for generating adversarial examples, so it is generalized against new attacks. Both well-trained models are employed to detect adversarial examples. No more models need to be trained during detecting adversarial examples, therefore the \textbf{AEAE} is a fast adversarial detector.

In summary, the architecture of the \textbf{AEAE} detector has the following advantages: (i) Since any adversarial examples are not utilized in the training process, this detector does not target specific attacks; (ii) This detector does not modify the architectures or parameters of the neural network and therefore do not result in accuracy loss; (iii) An autoencoder plays two roles, so this detector is inexpensive; (iv) This detector is independent of neural networks and complementary to other defenses and detections.

\section{Experimental Evaluation}
\subsection{Setup}
We utilize CIFAR-10 and ImageNet datasets for the image classification task. The pre-trained DenseNet model is employed to classify CIFAR-10 and the pre-trained VGG-19 model is employed to classify ImageNet. Both models achieve classification performance competitively with state-of-the-art results. The DenseNet model achieves top-1 accuracy of 94.84\%, the VGG-19 model achieves top-1 accuracy of 71.34\%, and top-5 accuracy of 90.02\%.

We evaluate the \textbf{AEAE} detector on all of the attacks described in Section \uppercase\expandafter{\romannumeral2}-A and summarized in Table~\ref{tab_attack}. As listed in Table~\ref{tab_attack}, the perturbation level in adversarial examples is related to the attack type and control parameters. For each of the attacks, we randomly select some images and generate their adversarial examples with different control parameters. These successful adversarial examples are utilized to evaluate the performance of the \textbf{AEAE} detector.
\begin{table}[htb]
\centering
\caption{Evaluation of Attacks.}
\begin{tabular}{|c|c|c|c|c|c|}
\hline
\multirow{2}*{Dataset} & \multirow{2}*{Attacks} & \multirow{2}*{Parameter} & \multicolumn{3}{c|}{Average $L_p$}  \\ \cline{4-6}
        &         &           &    $L_0$    &   $L_2$  &   $L_{\infty}$   \\
\hline
\multirow{15}*{CIFAR-10} & \multirow{3}*{FGSM} & $\epsilon = 0.1 $ & 0.992 & 5.405 & 0.102 \\
    & & $\epsilon = 0.2 $ & 0.993 & 10.551 & 0.200 \\
    & & $\epsilon = 0.3 $ & 0.993 & 15.366 & 0.302 \\
\cline{2-6}
& \multirow{3}*{BIM} & $\epsilon = 0.1 $ & 0.759  & 3.184  & 0.102 \\
    & & $\epsilon = 0.2 $ & 0.756  & 3.793  & 0.200 \\
    & & $\epsilon = 0.3 $ & 0.756  & 3.793  & 0.200\\
\cline{2-6}
& \multirow{3}*{PGD} & $\epsilon = 0.1 $ & 0.997 & 3.804  & 0.102 \\
    & & $\epsilon = 0.2 $ & 0.998 & 6.503  & 0.200 \\
    & & $\epsilon = 0.3 $ & 0.998 & 9.154  & 0.302 \\
\cline{2-6}
& DeepFool  & -  &  0.991  &  0.232  &  0.027 \\
\cline{2-6}
& \multirow{4}*{C\&W $L_2$} & $k = 0.0$ & 0.530 & 0.164  & 0.007 \\
    & & $k = 0.5$ & 0.425 & 0.184  & 0.018 \\
    & & $k = 1.0$ & 0.383 & 0.166  & 0.017 \\
    & & $k = 1.5$ & 0.347 & 0.157  & 0.017 \\
\hline
\multirow{15}*{ImageNet} & \multirow{3}*{FGSM} & $\epsilon = 0.1 $ & 0.988 & 37.618 & 0.102 \\
    & & $\epsilon = 0.2 $ & 0.988 & 73.084 & 0.200 \\
    & & $\epsilon = 0.3 $  & 0.988  &  106.119  & 0.302 \\
\cline{2-6}
& \multirow{3}*{BIM} & $\epsilon = 0.1 $ & 0.754  & 21.781  & 0.102 \\
    & & $\epsilon = 0.2 $ & 0.753  & 25.432  & 0.200 \\
    & & $\epsilon = 0.3 $ & 0.753  & 25.432  & 0.200 \\
\cline{2-6}
& \multirow{3}*{PGD} & $\epsilon = 0.1 $ & 0.995  & 26.888  & 0.102 \\
    & & $\epsilon = 0.2 $ & 0.996 & 47.050  & 0.200 \\
    & & $\epsilon = 0.3 $ & 0.996 & 65.578  & 0.302 \\
\cline{2-6}
& DeepFool  & -  &  0.977 &  0.833  &  0.031  \\
\cline{2-6}
& \multirow{4}*{C\&W $L_2$} & $k = 0.0$ & 0.404 & 0.950  & 0.004 \\
    & & $k = 0.5$ & 0.300 & 0.835  & 0.013 \\
    & & $k = 1.0$ & 0.244 & 0.742  & 0.013 \\
    & & $k = 1.5$ & 0.216 & 0.701  & 0.016 \\
\hline
\end{tabular}
\label{tab_attack}
\end{table}

To evaluate the robustness of the \textbf{AEAE}, we assume that attackers have full access to the target DNN model, but no ability to influence this model. Attackers are allowed to generate adversarial examples in a white-box manner, but they are unaware that our detector has been deployed.

\subsection{Reconstructed Image}
\begin{figure}[!htb]
\centering
\includegraphics[width=0.96\columnwidth]{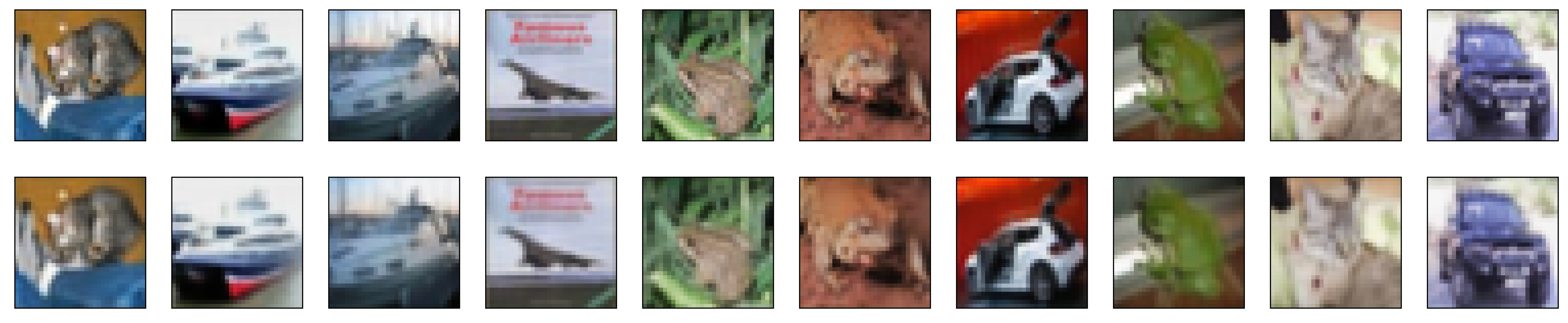}
\caption{Reconstructing CIFAR-10. Original images are listed in the first line, their reconstructed images are listed in the second line.}
\label{decoderCifar}
\end{figure}

\begin{figure}[!htb]
\centering
\subfloat{
    \includegraphics[width=0.18\columnwidth]{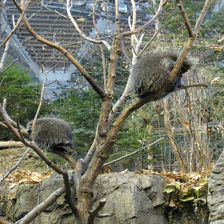}
}
\subfloat{
    \includegraphics[width=0.18\columnwidth]{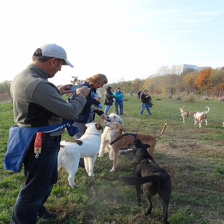}
}
\subfloat{
    \includegraphics[width=0.18\columnwidth]{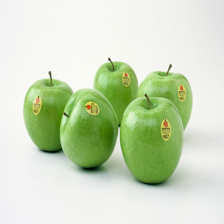}
}
\subfloat{
    \includegraphics[width=0.18\columnwidth]{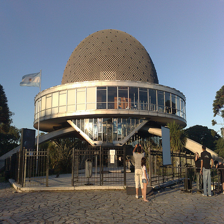}
}
\subfloat{
    \includegraphics[width=0.18\columnwidth]{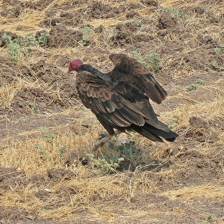}
}

\subfloat{
    \includegraphics[width=0.18\columnwidth]{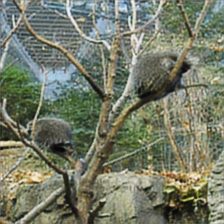}
}
\subfloat{
    \includegraphics[width=0.18\columnwidth]{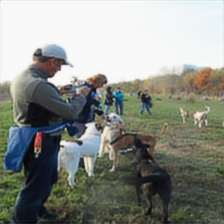}
}
\subfloat{
    \includegraphics[width=0.18\columnwidth]{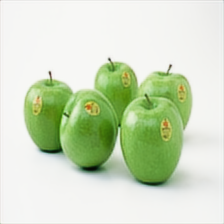}
}
\subfloat{
    \includegraphics[width=0.18\columnwidth]{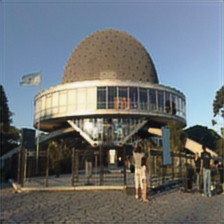}
}
\subfloat{
    \includegraphics[width=0.18\columnwidth]{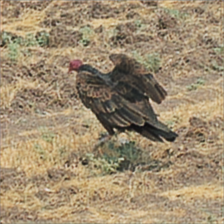}
}
\caption{Reconstructing ImageNet. Original images are listed in the first line, their reconstructed images are listed in the second line.}
\label{decoderImageNet}
\end{figure}

The $AEAE$ detector uses an autoencoder to learn features of benign images and filter out noise. This autoencoder should produce a small reconstruction error for the benign input to distinguish the adversarial one. We randomly choose 10 images from the CIFAR-10 dataset and 5 images from the ImageNet dataset and input them to the well-trained autoencoders. The original images and their reconstructed version are shown in Figure~\ref{decoderCifar} and Figure~\ref{decoderImageNet}.

In Figure~\ref{decoderCifar} and Figure~\ref{decoderImageNet}, original images are listed in the first line and their reconstructed images are listed in the second line. There is a high visual similarity between original images and reconstructed images. We calculate the average value of the reconstruction error. For the CIFAR-10 dataset, the average reconstruction error $MSE = 0.00051$; for the ImageNet dataset, the average reconstruction error $MSE = 0.00128$. The results show that our autoencoders produce small reconstruction errors and have good reconstruction ability.

\subsection{Overall Performance}
We evaluate the \textbf{AEAE} from the recall rate, precision rate, and the F1 score. The precision is intuitively the ability of the detector not to label as positive a sample that is negative. The recall is intuitively the ability of the classifier to find all the positive samples. The $F1$ score~\cite{LiangB} is a harmonic mean of the precision and recall. The recall, precision, and the F1 score are defined as follows,
\begin{equation}
\begin{gathered}
\text { Recall }=\frac{TP}{TP+FN} \\
\text { Precision }=\frac{TP}{TP+FP} \\
F1=2 * \frac{\text { Recall } * \text { Precision }}{\text { Recall }+\text { Precision }}
\end{gathered}
\label{metric}
\end{equation}
where $TP$ is the number of correctly detected adversarial examples (true positives), $FN$ is the number of adversarial examples that survive from the detector (false negatives), and $FP$ is the number of benign examples that are detected as adversarial examples (false positives). By taking their harmonic mean, $F1$ combines the precision and recall into a single metric to measure the overall detection performance. The highest possible value of an $F1$ is 1.0, indicating that all adversarial examples are detected without introducing any false positives.

In order to make a comparison with other detectors, we employ TPR (True Positive Rate) and FPR (False Positive Rate) to measure the performance on adversarial and benign examples, where TPR = Recall and FPR is defined as follows,
\begin{equation}
\text {FPR} = \frac{FP}{FP + TN}
\label{fpr}
\end{equation}
where $TN$ is the number of correctly detected benign examples. $FP + TN$ is the total number of benign examples. FPR (also known as the false alarm ratio) is employed to reflect detection ability on benign examples. The lowest possible value of an FPR is 0, indicating that all benign examples are correctly identified.

\subsubsection{Results on CIFAR-10}
We test the \textbf{AEAE} against FGSM, BIM, PGD, DeepFool, and C\&W attacks with multiple perturbation levels on CIFAR-10. The test images consist of 100 benign images and their successful adversarial examples generated by the above five types of attacks.

As listed in Table~\ref{tab_accCifar}, our detector achieves a recall of 93.95\% and a precision of 88.93\% against all adversarial attacks mentioned, the F1 score is up to 91.50\%. The \textbf{AEAE} shows a recall of up to 100\% in detecting adversarial examples generated by FGSM, BIM, and PGD attacks. These adversarial examples can be detected at a high detection rate even under different perturbation levels. For the DeepFool and the C\&W attacks with different perturbation levels, the \textbf{AEAE} achieves the average F1 scores of up to 85\% and shows robust detection ability. Our detector can detect adversarial examples on CIFAR-10 with a high detection rate, even if the attacker utilizes multiple attack methods to add different degrees of perturbation.
\begin{table}[htb]
\centering
\caption{Performance on CIFAR-10.}
\begin{tabular}{|c|c|c|c|c|}
\hline
Attacks & Parameter & Recall & Precision & F1 score  \\
\hline
\multirow{3}*{FGSM} & $\epsilon = 0.1 $ & 100\% & 89.90\% & 94.68\% \\
    & $\epsilon = 0.2 $ & 100\% & 90.00\% & 94.74\% \\
    & $\epsilon = 0.3 $  & 100\%  &  90.20\%  & 94.85\% \\
\hline
\multirow{3}*{BIM} & $\epsilon = 0.1 $ & 100\%  & 90.00\%  & 94.74\% \\
    & $\epsilon = 0.2 $ & 100\%  & 90.00\%  & 94.74\% \\
    & $\epsilon = 0.3 $ & 100\%  & 90.00\%  & 94.74\% \\
\hline
\multirow{3}*{PGD} & $\epsilon = 0.1 $ & 100\%  & 90.10\%  & 94.79\% \\
    & $\epsilon = 0.2 $ & 100\% & 90.20\%  & 94.85\% \\
    & $\epsilon = 0.3 $ & 100\% & 90.46\%  & 95.02\% \\
\hline
DeepFool  & -  &  82.98\%   &  88.64\%   &  85.71\%  \\
\hline
\multirow{4}*{C\&W $L_2$} & $k = 0.0$ & 79.17\% & 87.36\%  & 83.60\% \\
    & $k = 0.5$ & 85.42\% & 88.17\%  & 86.77\% \\
    & $k = 1.0$ & 83.33\% & 87.91\%  & 85.56\% \\
    & $k = 1.5$ & 84.38\% & 88.04\%  & 86.17\% \\
\hline
\multicolumn{2}{|c|}{Average} & 93.95\% & 88.93\%  & 91.50\% \\
\hline
\end{tabular}
\label{tab_accCifar}
\end{table}

\subsubsection{Results on ImageNet}
Compared to CIFAR-10, ImageNet is more challenging for classifiers, a well-trained VGG-19 model achieves top-1 accuracy of 66.10\%. We test the \textbf{AEAE} against FGSM, BIM, PGD, DeepFool, and C\&W attacks with different control parameters on ImageNet. The test images consist of 50 benign images and their successful adversarial examples generated by the above five types of attacks.

As listed in Table~\ref{tab_accImageNet}, our detector achieves an average recall of 81.64\% and an average precision of 70.51\%, the F1 score is 75.23\%. The \textbf{AEAE} seems to perform less successfully on ImageNet than on CIFAR-10. In particular, the \textbf{AEAE} is not robust enough against the BIM and DeepFool attacks. Due to the difficulty of detection on a high-resolution dataset, only a few detectors were tested on ImageNet and show poor detection capability\cite{AldahdoohHFD22}. For example, FS\cite{Xu18} respectively achieves 43\% and 64\% of detection rates for FGSM and BIM attacks. MagNet\cite{Meng17} has proven to be poor at detecting ImageNet images. Compared with these methods, our detector achieves a significant improvement in detecting high-resolution images.
\begin{table}[htb]
\caption{Performance on ImageNet.}
\centering
\begin{tabular}{|c|c|c|c|c|}
\hline
Attacks & Parameter & Recall & Precision & F1 score  \\
\hline
\multirow{3}*{FGSM} & $\epsilon = 0.1 $ & 100\% & 77.78\% & 87.50\% \\
    & $\epsilon = 0.2 $  & 100\%  &  78.13\%  & 87.72\% \\
    & $\epsilon = 0.3 $  & 100\%  &  78.13\%  & 87.72\% \\
\hline
\multirow{3}*{BIM} & $\epsilon = 0.1 $ & 64\%  & 68.00\%  & 66.41\% \\
    & $\epsilon = 0.2 $ & 66\%  & 70.21\%  & 68.04\% \\
    & $\epsilon = 0.3 $ & 66\%  & 70.21\%  & 68.04\% \\
\hline
\multirow{3}*{PGD} & $\epsilon = 0.1 $ & 98\%  & 77.78\% & 86.73\% \\
    & $\epsilon = 0.2 $ & 100\% & 78.13\% & 87.72\% \\
    & $\epsilon = 0.3 $ & 100\% & 78.13\% & 87.72\% \\
\hline
DeepFool  & -  &  61\%   &  68.19\%   &  64.52\%  \\
\hline
\multirow{4}*{C\&W $L_2$} & $k = 0.0$ & 84\% & 64.62\%  & 73.04\% \\
    & $k = 0.5$ & 70\% & 60.34\%  & 64.81\% \\
    & $k = 1.0$ & 70\% & 60.34\%  & 64.81\% \\
    & $k = 1.5$ & 60\% & 56.60\%  & 58.25\% \\
\hline
\multicolumn{2}{|c|}{Average} & 81.64\% & 70.51\%  & 75.23\% \\
\hline
\end{tabular}
\label{tab_accImageNet}
\end{table}

\subsubsection{Comparison}
We compare the \textbf{AEAE} detector with existing state-of-the-art methods. These methods include KD+BU\cite{FeinmanCSG17}, LID\cite{Ma0WEWSSHB18}, NSS\cite{KherchoucheFHD20}, NIC\cite{MaLTL019}, FS\cite{Xu18}, MagNet\cite{Meng17}. They are classified as supervised and supervised detectors and are mostly published in top-level conferences, such as CCS, NDSS, ICLR, IJCNN, etc. The comparison results are listed in Table~\ref{AEAEcompareAcc}. Comparing the two types of detectors, we find that the supervised detector usually has better performance than the unsupervised detector. This is in line with the fact that the supervised detector can learn features from adversarial examples to improve detection performance. However, there is a hidden danger that the supervised detector could have a poor detection rate for new types of adversarial examples. Since the training phase does not require any adversarial samples, the \emph{AEAE} falls into the unsupervised category and thus can avoid this hidden danger.

\begin{table*}[htb]
\caption{Comparison of Detection Capability}
\centering
\begin{tabular}{|c|c|c|c|c|c|c|c|c|c|c|c|c|}
\hline
\multirow{2}*{Detection type}&\multirow{2}*{Methods}& \multirow{2}*{Evaluation}&\multicolumn{2}{c|}{FGSM}&\multicolumn{2}{c|}{BIM}&\multicolumn{2}{c|}{PGD}&\multirow{2}*{DeepFool}&\multicolumn{2}{c|}{C\&W $L_2$}&\multirow{2}*{Average}\\ \cline{4-9} \cline{11-12}
                       &                   &                        &$\epsilon=8$&$\epsilon=16$&$\epsilon=8$&$\epsilon=16$&$\epsilon=8$& $\epsilon=16$ &  &$\epsilon=8$&$\epsilon=16$ & \\
\hline
\multirow{6}*{Supervised} & \multirow{2}*{KD+BU}  & TPR & 35.03\% & 33.23\% & 84.47\% & 99.55\% & 92.27\% & 99.89\% & 54.02\% & 37.29\%  & 29.69\% & 62.82\% \\
                      &                           & FPR & 7.3\%   & 4.5\%   & 2.18\%  & 0.07\%  & 0.96\%  & 0  &  1.44\% & 6.34\%   & 3.95\%  & 2.97\%  \\
                      & \multirow{2}*{LID}        & TPR & 53.0\%  & 81.23\% & 88.05\% & 98.55\% & 94.39\% & 99.22\% & 63.57\% & 44.59\%  & 65.46\% & 69.34\% \\
                      &                           & FPR & 3.84\%  & 1.44\%  & 3.65\%  & 0.44\%  & 1.81\%  & 0.26\% & 6.12\% & 15.01\%  & 19.25\% & 5.76\%  \\
                      & \multirow{2}*{NSS}        & TPR & 87.59\% & 99.94\% & 52.16\% & 87.74\% & 57.06\% & 93.24\% & 50.15\% & 40.94\%  & 65.12\% & 70.43\% \\
                      &                           & FPR & 6.56\%  & 6.56\%  & 6.56\%  & 6.56\%  & 6.56\%  & 6.56\% &  6.56\% & 6.56\%   & 6.56\%  & 6.56\%  \\
\hline
\multirow{10}*{Unsupervised}& \multirow{2}*{AEAE} & TPR & 100\%   & 100\%   & 100\%   & 100\%  & 100\%  & 100\% & 85.56\% & 84.21\% & 81.36\% & \textbf{95.57\%} \\
                      &                           & FPR & 10.0\%  & 10.0\%  & 10.0\%  & 10.0\% & 10.0\% & 10.0\% & 10.0\% & 10.0\%  & 10.0\%  & 10.0\% \\
                      & \multirow{2}*{NIC}        & TPR & 43.64\% & 58.48\% & 99.95\% & 100\%  & 100\%  & 100\% &  84.91\% & 75.18\% & 71.39\% & 81.51\% \\
                      &                           & FPR & 10.08\% & 10.08\% & 10.08\% & 10.08\%& 10.08\%& 10.08\% & 10.08\% & 10.08\% & 10.08\% & 10.08\% \\
                      & \multirow{2}*{FS}         & TPR & 29.33\% & 35.34\% & 8.74\%  & 0.34\% & 8.2\%  & 0.2\%  &  39.18\% & 68.33\% & 44.28\% & 25.99\% \\
                      &                           & FPR & 5.07\%  & 5.07\%  & 5.07\%  & 5.07\% & 5.07\% & 5.07\% &  5.07\% & 5.07\%  & 5.07\%  & 5.07\%  \\
                      & \multirow{2}*{MagNet}     & TPR & 0.72\%  & 3.11\%  & 0.56\%  & 0.69\% & 0.57\% & 0.66\% &  57.33\% & 0.61\%  & 0.44\%  & 7.19\%  \\
                      &                           & FPR & 0.77\%  & 0.77\%  & 0.77\%  & 0.77\% & 0.77\% & 0.77\% &  0.77\% & 0.77\%  & 0.77\%  & \textbf{0.77\%}  \\
\hline
\end{tabular}
\label{AEAEcompareAcc}
\end{table*}

 As listed in Table~\ref{AEAEcompareAcc}, the \emph{AEAE} achieves the average TPR of 95.70\%, which is the highest value when comparing both supervised and unsupervised detectors. The average TPR value of \emph{AEAE} detector is 25\% higher than that of the supervised NSS\cite{KherchoucheFHD20} and 14\% higher than that of the unsupervised NIC\cite{MaLTL019}. This result indicates that the \emph{AEAE} can detect adversarial examples with a high detection rate. In terms of FPR, the \emph{AEAE} reaches a relatively high value, which indicates that our detector has higher false alarms than most other detectors. Compared with NIC\cite{MaLTL019}, we find that the AEAE is obviously superior in both TPR rate and FPR, ensuring a high detection rate and a relatively low false alarm rate. In addition, we also observe that the \emph{AEAE} is not sensitive to perturbation levels and attack methods, which indicates that our detector is able to maintain stable performance for multiple attack methods with different perturbation levels.

\subsection{Case Study on PGD and C\&W Attacks}
Beyond considering the performance of our \emph{AEAE}, it is important to understand why this detector is effective against adversarial examples with different perturbation levels. To answer this question, we carry out a case study on both PGD and C\&W attacks against CIFAR-10.

In our detector \emph{AEAE}, we construct an isolation forest model and train it on a two-tuple feature dataset $D_{bf}=\left\{(MSE_1, KL_1),(MSE_2, KL_2),\dots\right\}$, where $MSE$ is the reconstruction error and $KL$ is prediction distance between benign images and their reconstructed versions. When a two-tuple $(MSE, KL)$ value from the unknown image is considered an outlier, this image is judged by the \emph{AEAE} to be an adversarial example.

\begin{figure}[htb]
\centering
\includegraphics[height=2.2in]{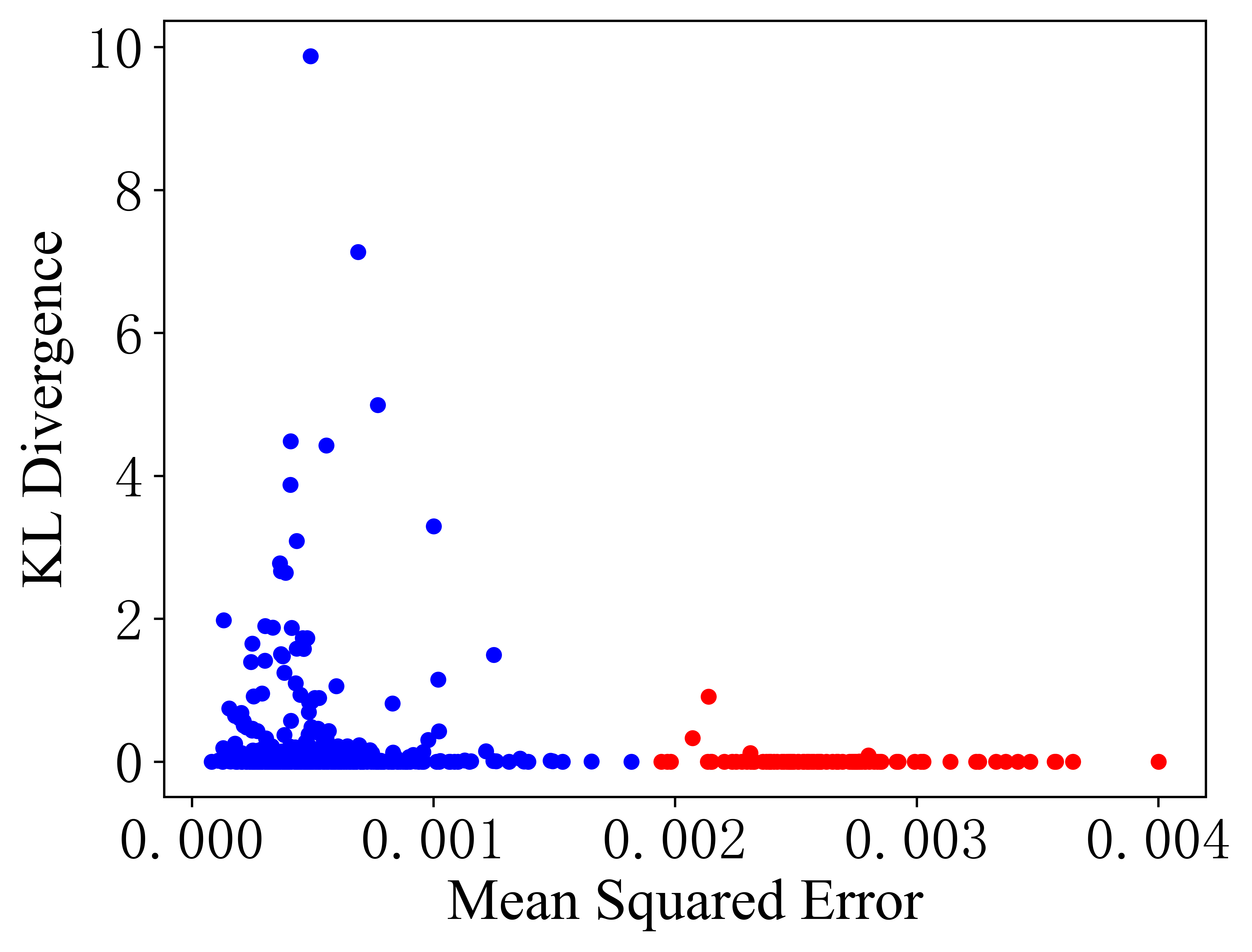}
\caption{Case study on the PGD attack. Blue dots represent benign features, red dots represent adversarial features.}
\label{case_PGD}
\end{figure}

\begin{figure}[htb]
\centering
\includegraphics[height=2.2in]{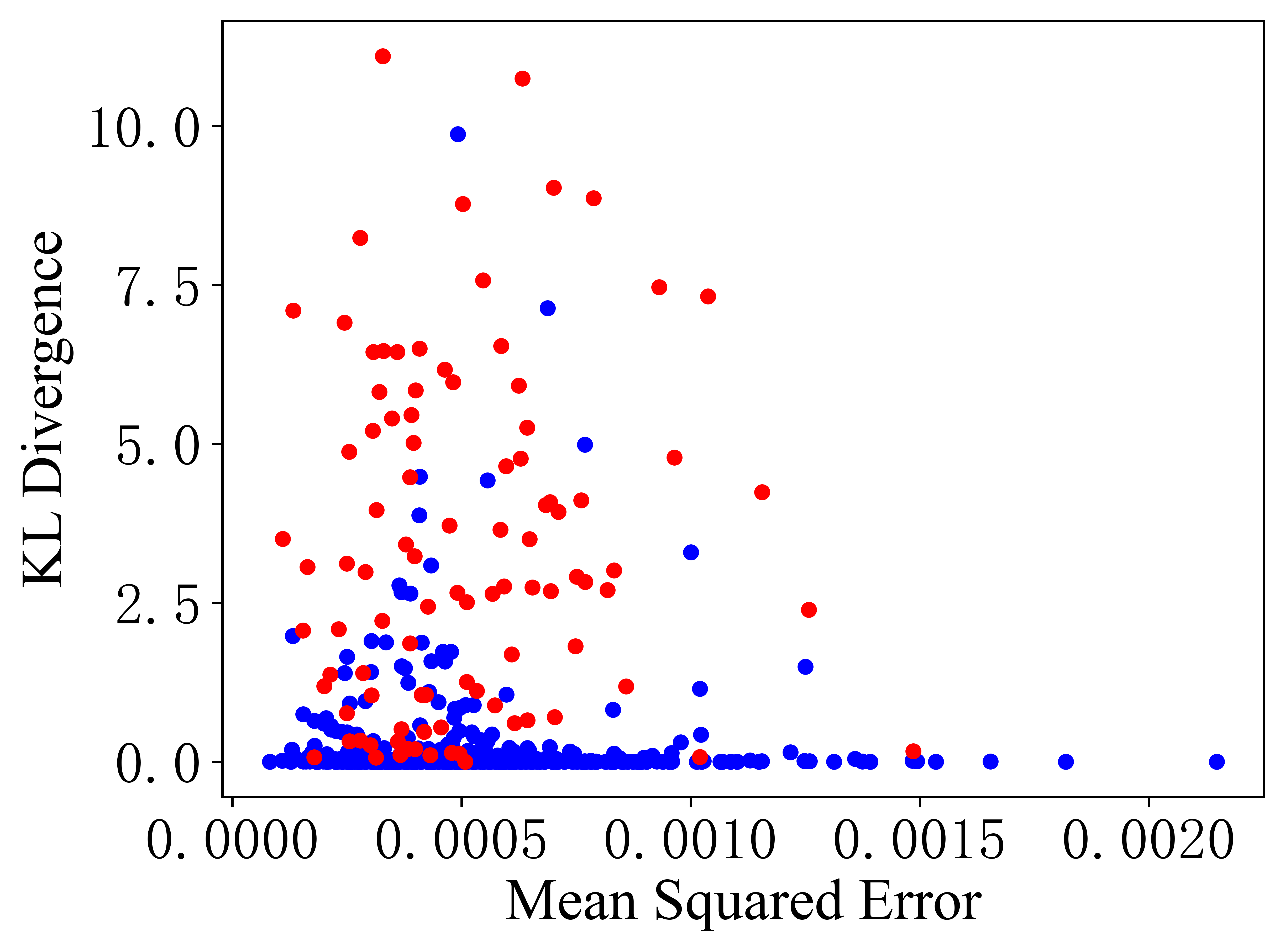}
\caption{Case study on the C\&W attack. Blue dots represent benign features, red dots represent adversarial features.}
\label{case_CW}
\end{figure}

We draw the scatter plot of $MSE$ vs. $KL$ from benign images and their adversarial examples. In Figure~\ref{case_PGD} and Figure~\ref{case_CW}, these two-tuples from benign images are marked in blue dots, and these two-tuples from adversarial images are marked in red dots. Figure~\ref{case_PGD} shows the scatter plot from benign images and adversarial images generated by the PGD attack with perturbation level $\epsilon = 0.1 $. Figure~\ref{case_CW} shows the scatter plot from benign images and adversarial images generated by the C\&W attack with perturbation level $k = 1.5$.

As shown in Figure~\ref{case_PGD}, all the red dots are to the right away from the blue dots, indicating that the PGD attack produces $MSE$ values outside the normal range. When we set an appropriate $MSE$ threshold, all adversarial examples generated by the PGD attack could be effectively detected. In our detector, the isolation forest learns the benign feature and forms an $MSE$-bound. If an $MSE$ value from an unknown image is outside this bound, our detector judges this image to be adversarial.

In Figure~\ref{case_CW}, the red dots are difficult to be separated from the blue dots by the $MSE$-bound. However, most of the red dots are above the blue dots, indicating that the C\&W attack produces larger $KL$ values than those from benign images. By learning the benign feature, our detector forms a $KL$-bound, which helps us to detect most adversarial examples generated by the C\&W attack.

By comparing Figure~\ref{case_PGD} and Figure~\ref{case_CW}, we find that the $MSE$ or $KL$ values alone are not guaranteed to be effective against any type of adversarial attack. In Figure~\ref{case_PGD}, the $KL$ value of benign images and PGD adversarial examples are both small and do not show a significant difference. We can separate them according to the $MSE$ value instead of the $KL$ value. Similarly, in Figure~\ref{case_CW}, the $MSE$ value of benign images and C\&W adversarial examples are both small. We can only distinguish them according to the $KL$ value. Both PGD and C\&W attacks usually produce an outlier of the $MSE$ or $KL$ values. Therefore, we construct a two-tuple feature $(MSE, KL)$ and train an isolation forest. This model can help us to detect adversarial examples with both significant and slight perturbations.

\subsection{Complexity}
The main price that the unsupervised detector pay is the overhead. Most of the unsupervised detection models train extra models to help the baseline model to detect the adversarial example\cite{AldahdoohHFD22}. The extra models need an additional storage space to be stored which might not be applicable to some devices and systems. The lightweight detection model has a wider range of application scenarios.

The \emph{AEAE} detector consists of a shallow autoencoder and an isolation forest model. This autoencoder includes 3 convolution layers, 1 max pooling layer, and 1 up sampling layer, with a total parameter of 40,451. The isolation forest model includes 100 binary trees, and each tree has no more than 400 nodes.

In the training phase, autoencoders are respectively trained on 50,000 images from the CIFAR-10 dataset and 1,000 images from the ImageNet dataset. Isolation forest models are trained on 400 two-tuple features from benign images. The inference phase is divided into two steps. In step (1), the well-trained autoencoder performs an inference to produce a reconstruction error and a prediction distance. These two values are constructed into a two-tuple feature. In step (2), This two-tuple feature is entered into the well-trained isolation forest model. If this model considers the two-tuple feature to be an outlier, this image would be judged as an adversarial example.

We calculate the model size and the inference time in two steps. As listed in Table~\ref{AEAEComplexity}, the size of the \emph{AEAE} detector is 503 KB and the running time for inferring a CIFAR-10 image is only 0.1974 seconds. Especially in step (2), the isolate forest algorithm performs effectively to detect outliers in small storage space. The results show that the $AEAE$ requires low storage overhead and has high detection efficiency.
\begin{table}[!htb]
\caption{Inference time and model size}
\centering
\begin{tabular}{ccc}
\hline
Step & size (KB)  & time (s)  \\
\hline
(1)  & 499 & 0.1971   \\
(2)  & 4   & 0.0003   \\
All  & 503 & 0.1974   \\
\hline
\end{tabular}
\label{AEAEComplexity}
\end{table}

In Table~\ref{AEAEcomparetime}, we compare the $AEAE$ with other unsupervised detectors, including NIC\cite{MaLTL019}, FS\cite{Xu18}, MagNet\cite{Meng17}, DNR\cite{SotgiuDMBFFR20} and SFAD\cite{AldahdoohHD22}. We evaluate the complexity(CM), overhead(OV) and inference time (INF) performance for each detector in 3-star ranking, where $\star = $ low, $\star\star = $ middle, $\star\star\star = $ high. As listed in Table~\ref{AEAEcomparetime}, the $AEAE$ detector has lower complexity and overhead, and higher inference efficiency than other unsupervised detectors. Thus, we can claim that the \emph{AEAE} is a lightweight unsupervised detector against adversarial examples.
\begin{table}[!htb]
\caption{Inference time and model size}
\centering
\begin{tabular}{ccccccc}
\hline
Evaluation & AEAE    & NIC &  FS   & MagNet   &  DNR   & SFAD \\
\hline
CM & $\star$ & $\star\star\star$ & $\star$     & $\star\star$  & $\star\star\star$  & $\star\star$ \\
OV & $\star$ & $\star\star\star$ & $\star$     & $\star\star$  & $\star\star\star$ & $\star\star\star$\\
INF& $\star$ & $\star\star\star$ & $\star\star$& $\star\star$  & $\star$           & $\star$\\
\hline
\end{tabular}
\label{AEAEcomparetime}
\end{table}

\section{Conclusion}
The unsupervised detection method does not target any type of adversarial attack and generally has better generalization. However, most unsupervised detectors need to train extra models to detect adversarial examples, resulting in additional storage space. In order to reduce the overhead, we propose a lightweight detector based on a shallow autoencoder and isolation forest. The autoencoder plays two roles. The first role learns the manifold of benign examples to detect significantly perturbed adversarial examples. The second role filters out the noise in the input to help to detect slightly perturbed adversarial examples. To cover both cases, we train an isolation forest on benign features as an alarm model. Since the autoencoder and the isolation forest are trained on benign data, the $AEAE$ is an unsupervised adversarial detector. We test the performance of the $AEAE$ on CIFAR-10 and ImageNet datasets. The results show that the $AEAE$ is more robust against adversarial examples with varying degrees of perturbation than other state-of-the-art detectors. The proposed detector only occupies a small amount of storage space and can efficiently detect adversarial examples.

\bibliographystyle{IEEEtran}
\bibliography{aeae}

\end{document}